\documentclass[sigconf]{aamas}  % do not change this line!
\renewcommand\footnotetextcopyrightpermission[1]{} % removes footnote with conference information in first column
\usepackage{algorithm}
\usepackage{filecontents}
\usepackage{algpseudocode}
\usepackage{caption}
\usepackage{subcaption}
\usepackage{amsfonts}
\usepackage{amsmath}
\usepackage{amssymb}
\usepackage{mathtools}
\usepackage{amsthm}
\usepackage{fmtcount} % for \ordinalnum
\usepackage{array,multirow}
\usepackage{xcolor}
\usepackage{balance} 

\definecolor{pastelgreen}{rgb}{0.01, 0.75, 0.24}
\definecolor{red(pigment)}{rgb}{0.93, 0.11, 0.14}
\definecolor{bleudefrance}{rgb}{0.19, 0.55, 0.91}

\DeclareMathOperator*{\argmin}{arg\,min}
\newcolumntype{C}[1]{>{\arraybackslash}m{#1}}

\newtheorem{dfn}{Definition}

\renewcommand{\algorithmicrequire}{\textbf{Input:}}
\renewcommand{\algorithmicensure}{\textbf{Output:}}

\setcopyright{ifaamas}  % do not change this line!
\acmDOI{doi}  % do not change this line!
\acmISBN{}  % do not change this line!
\acmConference[AAMAS'19]{Proc.\@ of the 18th International Conference on Autonomous Agents and Multiagent Systems (AAMAS 2019), N.~Agmon, M.~E.~Taylor, E.~Elkind, M.~Veloso (eds.)}{May 2019}{Montreal, Canada}  % do not change this line!
\acmYear{2019}  % do not change this line!
\copyrightyear{2019}  % do not change this line!
\acmPrice{}  % do not change this line!

\begin{document}

\title{Explicable Planning as Minimizing Distance\\from Expected Behavior}

%\titlenote{Produces the permission block, and copyright information}

% AAMAS: as appropriate, uncomment one subtitle line; check the CFP
%\subtitle{Extended Abstract}
%\subtitle{Industrial Applications Track}
%\subtitle{Socially Interactive Agents Track}
%\subtitle{Blue Sky Ideas Track}
%\subtitle{Engineering Multiagent Systems Track}
%\subtitle{Robotics Track}
%\subtitle{JAAMAS Track}
%\subtitle{Doctoral Mentoring Program}

%\subtitlenote{The full version of the author's guide is available as \texttt{acmart.pdf} document}

% AAMAS: submissions are anonymous for most tracks
\author{Anagha Kulkarni}
\affiliation{%
  \institution{Arizona State University}
}
\email{anaghak@asu.edu}

\author{Yantian Zha}
\affiliation{%
  \institution{Arizona State University}
}
\email{yzha3@asu.edu}

\author{Tathagata Chakraborti}
\affiliation{%
  \institution{IBM Research AI}
}
\email{tchakra2@ibm.com}

\author{Satya Gautam Vadlamudi}
\affiliation{%
  \institution{CreditVidya}
}
\email{gautam.vadlamudi@creditvidya.com}

\author{Yu Zhang}
\affiliation{%
  \institution{Arizona State University}
}
\email{yzhan442@asu.edu}

\author{Subbarao Kambhampati}
\affiliation{%
  \institution{Arizona State University}
}
\email{rao@asu.edu}

\begin{abstract} 
In order to have effective human-AI collaboration, it is necessary to address how the AI agent's behavior is being perceived by the humans-in-the-loop. When the agent's task plans are generated without such considerations, they may often demonstrate \emph{inexplicable behavior} from the human observer's point of view. This problem may arise due to the observer's partial or inaccurate understanding of the agent's planning model. This may have serious implications, from increased cognitive load to more serious concerns of safety around the physical agent. In this paper, we address this issue by modeling the notion of plan explicability as a function of the distance between a plan that the agent makes and the plan that the observer expects it to make. We achieve this, by learning a regression model over the plan distances and map them to the labeling scheme used by the human observers. We develop an anytime search algorithm that is guided by a heuristic derived from the learned model to come up with progressively explicable plans. We evaluate the effectiveness of our approach in a simulated autonomous car domain and a physical robot domain and also report results of the user studies performed in both the domains. 
\end{abstract}

%AAMAS: the ACM CCS are encouraged but optional within AAMAS papers


  % put your semicolon-separated keywords here!

\maketitle

%%%%%%%%%%%%%%%%%%%%%%%%%%%%%%%%%%%%%%%%%%%%%%%%%%%%%%%%%%%%%%%%%%%%%%%%%%%%%%%%%%%%%%%%%%%%%%%%%%%%%%%%%
%% start of main body of paper

\section{Introduction}

There is a growing interest in the applications that involve human-robot collaboration. 
An important challenge in such human-in-the-loop scenarios is to ensure that the robot's behavior is not just optimal with respect to its own model, but is also explicable and comprehensible to the humans in the loop. 
Without it, the robot runs the risk of increasing the cognitive load of humans which can result in reduced productivity, safety, and trust \cite{fan2008influence}. 
This mismatch between the robot's plans and the human's expectations of the robot's plans may arise because of the difference in the actual robot model $\mathcal{M^R}$, and the \emph{human's expectation of the robot model} $\mathcal{M^R_H}$.  
For example, consider a scenario with an autonomous car switching lanes on a highway. The autonomous car, in order to switch the lane, may make sharp and calculated moves, as opposed to gradually moving towards the other lane. These moves may well be optimal for the car due to its superior sensing and steering capabilities. Nevertheless, a passenger sitting inside may perceive this as dangerous and reckless behavior, in as much as they might be ascribing to the car driving abilities that they themselves have. 

\begin{figure}[t!]
\centering 
\includegraphics[width=\columnwidth]{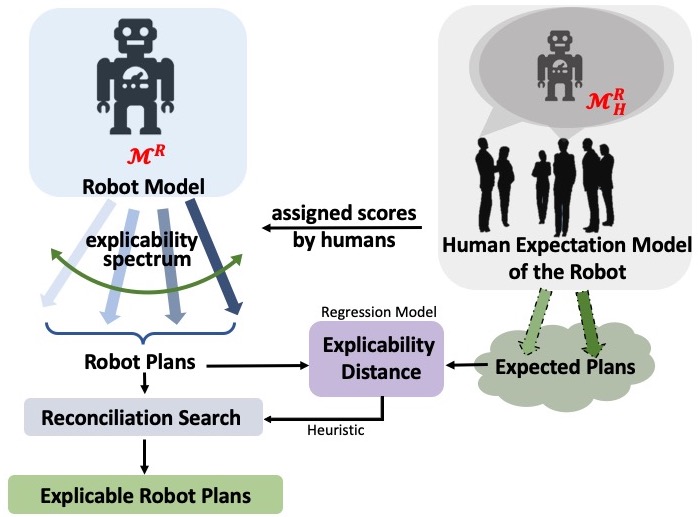}
\caption{Schematic diagram of the setting: Here a regression model called explicability distance is learned to fit plan scores assigned by humans to plan distances between the robot's and the human's expected plans. This gives a heuristic for computing explicable plans, which is used by the process of reconciliation search introduced in the paper.}
\label{fig:architecture}
\end{figure}

In order to avoid inexplicable behavior, the robot has to take the human mental model into account and compute the plans that align with this model. As long as the robot's behavior is aligned with the human mental model, the human can make sense of it. Therefore, the objective of explicable planning is to generate robot plans that not only minimize the cost of the plan, but also the distance between the plans produced by$\mathcal{M^R}$ and $\mathcal{M^R_H}$. Of course, an immediate question is, if $\mathcal{M^R_H}$ is available to the robot, why is $\mathcal{M^R}$ required in the plan generation process at all? We note that this is a necessary component since the human mental model might entail plans that are not even feasible for the robot or are prohibitively expensive, and can thus at best serve as a guide, and not an oracle, to the explicable plan generation process. Therefore, instead of using $\mathcal{M^R_H}$ directly, the robot can use $\mathcal{M^R_H}$ as a guide to compute plans that reduce the distance with human's expected plans. In settings where the objective is to minimize the cognitive load on the human or minimize the cost of explicit communication of explanations \cite{explain,pat}, the computation of explicable plans can be crucial. Also, settings where the observers are not necessarily experts in the domain and tend to have noisy or incomplete understanding of robot behavior, explicable plans can be useful for engendering trust. Some example application domains for this setting could be mission planning, urban search and rescue, and household domains.

An important consideration in the computation of explicable behavior is access to the human mental model, $\mathcal{M^R_H}$. In many domains, such as in factory scenarios, mission planning or household), there is generally a clear expectation of how a task should be performed. In such cases, $\mathcal{M^R_H}$ can be constructed following the norms or protocols that are relevant to that domain. Most deployed products make use of inbuilt models of user expectations in some form or the other. 
%Building such models, of course, require interactions with users of that domain. 
In this setting, we hypothesize that the plan distances \cite{srivastava2007domain,nguyen-partialp-2012} can quantify the distance between the robot plan $\pi_{\mathcal{M^R}}$ and the expected plans $\pi_{\mathcal{M^R_H}}$ from $\mathcal{M^R_H}$. The domain modeler constructs $\mathcal{M^R_H}$, which is then used to generate expected plans. Then we compute the distance between plans from $\mathcal{M^R}$ and $\mathcal{M^R_H}$. The test subjects provide explicability assessments (scores reflecting explicability) for robot plans. Then the scores are mapped to the precomputed distances and a regression model of the explicability distance is learned. The plan generation process uses this
learned explicability distance as a heuristic to guide the search. This process is illustrated in Figure \ref{fig:architecture}.

In the following section, we will formally define the
plan explicability problem as introduced above, and propose an anytime search approach -- 
{\em reconciliation search} -- that can generate solution with varying levels of explicability 
given the human mental model.
We also establish of the effectiveness of our approach by empirical evaluations and user studies for two domains: a simulated robot car domain and a physical robot delivery domain.

\section{Explicable Planning Problem}

\subsection{Classical Planning}

A classical planning problem can be defined as a tuple $\mathcal{P}= \langle \mathcal{M}, \mathcal{I}, \mathcal{G} \rangle $, where $\mathcal{M} = \langle \mathcal{F}, \mathcal{A} \rangle$ is the domain model (that consists of a finite set $\mathcal{F}$ of fluents that define the state of the world and a set of operators or actions $\mathcal{A}$), and $\mathcal{I} \subseteq \mathcal{F}$ and $\mathcal{G} \subseteq \mathcal{F}$ are the initial and goal states of the problem respectively. 
Each action $a \in \mathcal{A}$ is a tuple of the form $\langle pre(a), eff(a), c(a) \rangle$ where $c(a)$ denotes the cost of an action, $pre(a) \subseteq \mathcal{F}$ is the set of preconditions for the action $a$ and $eff(a) \subseteq \mathcal{F}$ is the set of the effects. The solution to the planning problem is a \emph{plan} or a sequence of actions $\pi = \langle a_1, a_2, \ldots, a_n \rangle$ such that starting from the initial state, sequentially executing the actions lands the agent in the goal state, i.e. $\Gamma_\mathcal{M}(\mathcal{I}, \pi) \models \mathcal{G}$ where $\Gamma_\mathcal{M}(\cdot)$ is the transition function defined for the domain. The cost of the plan, denoted as $c(\pi)$, is given by the summation of the cost of all the actions in the plan $\pi$, $c(\pi) = \sum_{a_i\in\pi}c(a_i)$. 

\subsection{Problem Definition}

The problem of explicable planning arises when the robot plan, $\pi_{\mathcal{M^R}}$, deviates from the human's expectation of that plan, $\pi_{\mathcal{M^R_H}}$. Here $\pi_{\mathcal{M^R}}$ is the robot's plan solution to the planning problem,  $\mathcal{P^R}= \langle \mathcal{M^R}, \mathcal{I^R}, \mathcal{G^R} \rangle $, where $\mathcal{M^R} = \langle \mathcal{F^R}, \mathcal{A^R} \rangle$; whereas, $\pi_{\mathcal{M^R_H}}$ is the plan solution considering the human mental model, such that, $\mathcal{P^R_H}= \langle \mathcal{M^R_H}, \mathcal{I^R}, \mathcal{G^R} \rangle $, where $\mathcal{M^R_H} = \langle \mathcal{F^R}, \mathcal{A^R_H} \rangle$. The differences in the human mental model can lead to different plan solutions. 

% \newpage

% We now formally define the explicable planning problem and the solution to it. 

\begin{dfn}
The \textbf{explicable planning problem} is defined as a tuple $\mathcal{P}_{EPP} = \langle \mathcal{M^R}, \mathcal{M^R_H}, \mathcal{I^R}, \mathcal{G^R} \rangle $, where, \\[-1ex]
\begin{itemize}
\item $\mathcal{M^R} = \langle \mathcal{F^R}, \mathcal{A^R} \rangle$ is the robot's domain model where $\mathcal{F^R}$ and $\mathcal{A^R}$ represent the actual set of fluents and robot actions \\[-1ex]
\item $\mathcal{M^R_H} = \langle \mathcal{F^R}, \mathcal{A^R_H} \rangle$ is the human's mental model of the robot model where $\mathcal{F^R}$ and $\mathcal{A^R_H}$ represent the set of fluents and actions that the human thinks are available to the robot \\[-1ex]
\item $\mathcal{I^R}$ is the initial state of the robot \\[-1ex]
\item $\mathcal{G^R}$ is the goal state of the robot \\[-1ex]
\end{itemize}
\end{dfn}

Here $\mathcal{A^R}$ and $\mathcal{A^R_H}$ represent that the action names, preconditions, effects and costs of the actions can be different. The initial state and the goal state are known to the human-in-the-loop. In order to compute the difference between the robot plan and the plan expected by a human, we need an evaluation function that computes the distance in terms of various aspects of the plan, like the action sequence, state sequence, etc. Let's denote such an evaluation function as $\delta^*$. The solution to an explicable planning problem is an explicable plan that achieves the goal and minimizes the model differences while also minimizing the cost of the plan. We now formally define it as follows:

\begin{dfn} \cite{exp-yz}
A plan is an \textbf{explicable plan}, $\pi^{*}_{\mathcal{M^R}}$, starting at $\mathcal{I^R}$ that achieves the goal $\mathcal{G^R}$, such that, $\displaystyle \argmin_{\pi_{\mathcal{M^R}}} c(\pi_{\mathcal{M^R}}) + \delta^*(\pi_{\mathcal{M^R}}, \pi_{\mathcal{M^R_H}}) $, where $\delta^*$ is an evaluation function. 
\end{dfn}

Since the evaluation function is not directly available to us, we learn an approximation of it using a combination of three plan distances measures. In the following section, we discuss our approach for learning the evaluation function and computing explicable plans.

\section{Proposed Methodology}

In this section, we quantify the explicability of the robot plans in terms of the \emph{plan distance} between the robot plan $\pi_{\mathcal{M^R}}$ and candidate expected plans $\pi_{\mathcal{M^R_H}}$ from $\mathcal{M^R_H}$. The outline of our approach is illustrated in Figure \ref{fig:architecture}. Given both the models $\mathcal{M^R}$ and $\mathcal{M^R_H}$ are obtained, our approach takes the following steps:

\vspace{5pt}
\begin{enumerate}
\item Firstly, the plan distances between the robot plans and the expected plans are computed using plan the three aforementioned plan distance measures. \\[-2ex]
\item The human subjects are asked to provide scores for each of the candidate robot plans by labeling each action in the plan with an explicable or inexplicable label. \\[-2ex]
\item Then the human explicability assessments (scores reflecting explicability) of candidate robot plans are mapped to the plan distance measures in form of regression model called explicability distance.  \\[-2ex]
\item The synthesis of explicable plans is achieved by modifying the \texttt{Fast-Downward} \cite{fastd} planner to incorporate an anytime search with explicability distance as the heuristic. This process results in incrementally more explicable plans. 
\end{enumerate}

% In the following subsections, we explain each of these steps in detail and formally define necessary terms.

\subsection{Background on Plan Distance Measures}

We now introduce three plan distances -- action, causal link and state sequence distances -- 
proposed in \cite{srivastava2007domain,nguyen-partialp-2012}, that we will reuse in this work
to capture the explicability distance between plans. 

\paragraph{Action distance}

We denote the set of unique actions in a plan $\pi$ as $A(\pi) = \{a~|~a\in\pi\}$.
Given the action sets $A(\pi_{\mathcal{M^R}})$ and $A(\pi^*_{\mathcal{M^R_H}})$ of two plans $\pi_{\mathcal{M^R}}$ and $\pi^*_{\mathcal{M^R_H}}$ respectively, the action distance is, 
\begin{equation} \label{eqn:eqn1}
\delta_A(\pi_{\mathcal{M^R}}, \pi^*_{\mathcal{M^R_H}}) = 1 - \frac{\lvert A(\pi_{\mathcal{M^R}}) \cap A(\pi^*_{\mathcal{M^R_H}}) \rvert}{\lvert A(\pi_{\mathcal{M^R}}) \cup A(\pi^*_{\mathcal{M^R_H}}) \rvert}
\end{equation} 
Here, two plans are similar (and hence their distance measure is smaller) if they contain same actions. Note that it does not consider the ordering of actions. 

\paragraph{Causal link distance}

A causal link represents a tuple of the form $\langle a_i, p_i, a_{i+1} \rangle$, where $p_{i}$ is a predicate variable that is produced as an effect of action $a_i$ and used as a precondition for the next action $a_{i+1}$. The causal link distance measure is represented using the causal link sets $Cl(\pi_{\mathcal{M^R}})$ and $Cl(\pi^*_{\mathcal{M^R_H}})$,
\begin{equation} \label{eqn:eqn2}
\delta_C(\pi_{\mathcal{M^R}}, \pi^*_{\mathcal{M^R_H}}) = 1 - \frac{\lvert Cl(\pi_{\mathcal{M^R}}) \cap Cl(\pi^*_{\mathcal{M^R_H}}) \rvert}{\lvert Cl(\pi_{\mathcal{M^R}}) \cup Cl(\pi^*_{\mathcal{M^R_H}}) \rvert}
\end{equation}

\paragraph{State sequence distance}

This distance measure finds the difference between sequences of the states. Given two state sequences $(s^R_0, \ldots, s^R_n)$ and $(s^H_0, \ldots, s^H_{n^{\prime}})$ for $\pi_{\mathcal{M^R}}$ and $\pi^*_{\mathcal{M^R_H}}$ respectively, where $n \geq n^{\prime}$ are the lengths of the plans, the state sequence distance is, \begin{equation} \label{eqn:eqn3}
\delta_S(\pi_{\mathcal{M^R}}, \pi^*_{\mathcal{M^R_H}}) = \frac{1}{n}  \big[ \  \sum_{k=1}^{n^{\prime}} d(s_k^R, s_k^H) + n - n^{\prime} \big] \ \
\end{equation} 
where,
\begin{equation} d(s_k^R, s_k^H) = 1 - \frac{\lvert s_k^{R} \cap s_k^{H} \rvert}{\lvert s_k^{R} \cup s_k^{H} \rvert} \end{equation} represents the distance between two states (where $s_k^{R}$ is overloaded to denote the set of predicate variables in state $s_k^{R}$). The first term measures the normalized difference between states up to the end of the shortest plan, while the second term, in the absence of a state to compare to, assigns maximum difference possible.

\subsection{Explicability Distance}

We start with a general formulation for capturing a measure of explicability of the robot's plans using plan distances. 
% present a general formulation in this section Since, without the model we do not know which plan distance is most relevant in capturing explicability, we present a general formulation in this section. 
%A more detailed formulation can be found in the subsequent section. 
A set of robot plans are scored by humans such that each action that follows the human's expectation in the context of the plan is scored 1 if explicable (0 otherwise). The plan score is then computed as the ratio of the number of explicable actions to the total plan length. 

\begin{dfn}
A set of plans is a \textbf{set of expected plans}, $\mathcal{E(P^R_H)}$, for the planning problem $\mathcal{P^R_H} = \langle \mathcal{M^R_H}, \mathcal{I^R}, \mathcal{G^R} \rangle$,  is a set of optimal cost plans that solve $\mathcal{P^R_H}$, $\mathcal{E(P^R_H)} = \{ \pi^{(i)}_{\mathcal{M^R_H}} | i = 1, \ldots, n\}$.
\end{dfn}

This set of expected plans consists of the plan solutions that the human expects the robot to compute. But these plans are not necessarily feasible in the robot model, $\mathcal{M^R}$.

\begin{dfn}
A \textbf{composite distance}, $\delta_{exp}$ is a distance between pair of two plans $\langle \pi_{\mathcal{M^R}}, \pi_{\mathcal{M^R_H}} \rangle$, such that, $\delta_{exp}(\pi_{\mathcal{M^R}}, \pi_{\mathcal{M^R_H}}) = || \delta_A(\pi_{\mathcal{M^R}}, \pi_{\mathcal{M^R_H}})  + \delta_C(\pi_{\mathcal{M^R}}, \pi_{\mathcal{M^R_H}}) + \delta_S(\pi_{\mathcal{M^R}}, \pi_{\mathcal{M^R_H}}) ||_2 $.
\end{dfn}

Among the set of expected plans, we compute a distance minimizing plan as follows: 

\begin{dfn}
A \textbf{distance minimizing plan}, ${\pi}^{*}_{\mathcal{M^R_H}}$, is a plan in $\mathcal{E(P^R_H)}$, such that for a robot plan, $\pi_{\mathcal{M^R}}$, the composite distance is minimized, i.e --
\begin{equation}
{\pi}^{*}_{\mathcal{M^R_H}} = 
\{ \pi_{\mathcal{M^R_H}} | ~ \displaystyle \argmin_{\pi_{\mathcal{M^R_H}}} \delta_{exp}(\pi_{\mathcal{M^R}}, \pi_{\mathcal{M^R_H}}) \}
\end{equation}
\end{dfn}

\begin{dfn}
\label{def:4}
An \textbf{explicability feature vector}, $\Delta$, is a three-dimensional vector, which is given with respect to a distance minimizing plan pair, $\langle \pi_{\mathcal{M^R}}, \pi^{*}_{\mathcal{M^R_H}} \rangle$, such that, \begin{equation}
\Delta= \langle \delta_A(\pi_{\mathcal{M^R}}, \pi^{*}_{\mathcal{M^R_H}}), \delta_C(\pi_{\mathcal{M^R}}, \pi^{*}_{\mathcal{M^R_H}}), \delta_S(\pi_{\mathcal{M^R}}, \pi^{*}_{\mathcal{M^R_H}}) \rangle^T 
\end{equation}
\end{dfn}

We approximate the evaluation function, $\delta^*$, using a combination of the three plan distance measures as follows: 

\begin{dfn}
The \textbf{explicability distance function}, \\ $\textit{Exp}(\pi_{\mathcal{M^R}}~/~\pi^{*}_{\mathcal{M^R_H}})$, is a regression function, \textit{f}, that fits the three plan distances to the total plan scores, with $b$ as the parameter vector, and $\Delta$ as the explicability feature vector, such that, 
\begin{equation}
\textit{Exp}(\pi_{\mathcal{M^R}}~/~\pi^{*}_{\mathcal{M^R_H}}) \approx \textit{f} (\Delta, b)
\end{equation}
\end{dfn}

A regression model is trained to learn the explicability assessment (total plan scores) of the users by mapping this assessment to the explicability feature vector which consists of plan distances for corresponding plans. 

\subsection{Plan Generation}

In this section, we present the details of our plan generation phase. We use the learned explicability distance function as a heuristic to guide our search towards explicable plans. 

\begin{algorithm}[tbp!]
\raggedright
\caption{Reconciliation Search}
\label{search1}
\algorithmicrequire{ $\mathcal{P_{EPP}} = \langle \mathbb{ \mathcal{M^R}, \mathcal{M^R_H}}, \mathcal{I^R}, \mathcal{G^R}\rangle$, $max\_cost$, and explicability distance function \textit{Exp}(.~,~.)}\newline
\algorithmicensure{ $\mathcal{E_{EPP}}$}
\begin{algorithmic}[1]
\State $\mathcal{E_{EPP}} \gets \emptyset $ \Comment{Explicable plan set}
\State $\textit{open} \gets \emptyset$  \Comment{Open list}
\State $\textit{closed} \gets \emptyset$  \Comment{Closed list}
\State $\textit{open}$.insert$(\mathcal{I}, 0, \inf)$
\While {$\textit{open} \neq \emptyset$}
\State  $\textit{n} \gets \textit{open}$.remove() \Comment{Node with highest $\textit{h}(\cdot)$}
\If {$\textit{n} \models \mathcal{G}$}
\State $\mathcal{E_{EPP}}$.insert($\pi \text{ s.t. } \Gamma_{\mathcal{M^R}}(\mathcal{I}, \pi) \models \textit{n} $)
\EndIf
\State $\textit{closed}$.insert($\textit{n}$)
\For {each $\textit{v} \in $ successors($\textit{n}$)}
\If {$\textit{v} \notin \textit{closed}$}
\If {$\textit{g(n)} + \textit{cost}(\textit{n}, \textit{v}) \leq \textit{max\_cost}$}
\State $\textit{open}$.insert($\textit{v}, \textit{g(n)} + \textit{cost}(\textit{n}, \textit{v}), \textit{h(v)}$)  
%\Comment{Plan $\pi_v$ from $\mathbb{I}$ to $v$}
\EndIf
\Else
\If {$\textit{h(n)} < \textit{h(v)}$}
\State $\textit{closed}$.remove($\textit{v}$)
\State $\textit{open}$.insert($\textit{v}, \textit{g(n)} + \textit{cost(n, v)}, \textit{h(v)}$)
\EndIf
\EndIf
\EndFor
\EndWhile
\State \Return $\mathcal{E_{EPP}}$ 
\end{algorithmic}
\end{algorithm}

\subsection*{Reconciliation Search}

The solution to an explicable planning problem $\mathcal{P_{EPP}} = \langle \mathcal{M^R}, $ $\mathcal{M^R_H}, \mathcal{I^R}, \mathcal{G^R} \rangle$ is the set $\mathcal{E_{EPP}}$ of explicable plans 
(with varying degrees of explicability) in $\mathcal{M^R}$.
This is found by performing ``reconciliation search'' (as detailed in Algorithm \ref{search1}). 

% Given an explicable planning problem $\mathcal{P_{EPP}} = \langle \mathcal{M^R}, \mathcal{M^R_H}, \mathcal{I^R}, \mathcal{G^R} \rangle$, the set of explicable plans in $\mathcal{M^R}$ are defined as, $\mathcal{E_{EPP}} = \{ \pi^{*(i)}_{\mathcal{M^R}} | i = 1, \ldots, m\}$, such that, each $\pi^{*(i)}_{\mathcal{M^R}}$ is found by performing ``reconciliation search'' (Algorithm \ref{search1}). Here each $\pi^{*(i)}_{\mathcal{M^R}}$ may have different degree of explicability. 

%In the following subsections, we discuss how this property affects our heuristic and how it is handled in the search process. 

\subsection*{Non-Monotonicity}
Since plan score is the fraction of explicable actions in a plan, it exhibits non-monotonicity. As a partial plan grows, a new action may contribute either positively or negatively to the plan score, thus making the explicability distance function non-monotonic. Consider that the goal of an autonomous car is to park itself in a parking spot on its left side. The car takes the left turn, parks and turns on its left indicator. Here the turning on of the left tail light after having parked is an inexplicable action. The first two actions are explicable to the human drivers and contribute positively to the explicability score of the plan but the last action has a negative impact and decreases the score. 

Due to non-monotonic nature of explicability distance, we cannot stop the search process after finding the first solution. Consider the following: if $\textit{e}_1$ is explicability distance of the first plan, then a node may exist in open list (set of unexpanded nodes) whose explicability distance is less than $\textit{e}_1$, which when expanded may result in a solution plan with explicability distance higher than $\textit{e}_1$. A greedy method that expands a node with the highest explicability score of the corresponding partial plan at each step is not guaranteed to find an optimal explicable plan (one of the plans with the highest explicability score) as its first solution.
Therefore, to handle the non-monotonic nature, we present a cost-bounded anytime greedy search algorithm called \emph{reconciliation search} that generates all the valid loopless candidate plans up to a given cost bound, and then progressively searches for plans with better explicability scores. 
The value of the heuristic $\textit{h(v)}$ in a particular state $\textit{v}$ encountered during search is based entirely on the explicability distance of the agent plan prefix $\pi_{\mathcal{M^R}}$ up to that state,
\begin{eqnarray} 
&\textit{h(v)}\  = \textit{Exp}(\pi_{\mathcal{M^R}}/\pi^{\prime}_{\mathcal{M^R_H}}) \nonumber \\
&\text{ s.t. } \Gamma_{\mathcal{M^R}}(\mathcal{I}, \pi_{\mathcal{M^R}}) \models v ~\text{and } \nonumber \\
&\Gamma_{\mathcal{M^R_H}}(\mathcal{I}, \pi^{\prime}_{\mathcal{M^R_H}}) \models v
\label{heu}
\end{eqnarray}

Note that, we assume that the same state space is reachable for computation of the plan prefix $\pi'$ from $\mathcal{I}$ to $\textit{v}$ in $\mathcal{M^R_H}$ (as per Equation \ref{heu}). 

We implement this search in the \texttt{Fast-Downward} planner. The approach is described in detail in Algorithm \ref{search1}. At each iteration of the algorithm, the plan prefix of the agent model is compared with the explicable trace $\pi^{\prime}_{\mathcal{M^R_H}}$ (these are the plans generated using $\mathcal{M^R_H}$ up to the current state in the search process) for the given problem. Using the computed distances, the explicability score for the candidate agent plans is predicted. The search algorithm then makes a locally optimal choice of states. We will make this clearer in the revised manuscript.
 We do not stop the search after generating the first solution, but instead, continue to find all the valid loopless candidate solutions within the given cost bound or until the state space is completely explored.

\section{Experimental Analysis}

Here we present the evaluation of our system in simulated autonomous car domain and physical robot delivery domain.

\subsection{Autonomous Car Domain}

\subsubsection*{Domain model}

There are a lot of social norms followed by human drivers which are usually above explicit laws. This can include normative behavior while changing lanes or during turn-taking at intersections. As such this car domain has emerged as a vibrant testbed for research \cite{sadigh2016planning,ijcai2017-664} in the HRI/AI community in recent times. In this work, we explore the topic of explicable behavior of an autonomous car in this domain, especially as it related to mental modeling of the humans in the loop.
In our autonomous car domain (modeled in PDDL), the autonomous car model $\mathcal{M^R}$ consists of lane and car objects as shown in Figure \ref{fig:car_sim}. The red car is the autonomous car in the experiments and all other cars are assumed to have human drivers. The car objects are associated with predicates defining the location of the car on a lane segment, status of left and right turn lights, whether the car is within the speed limit, the presence of a parked police car, and so on. The actions possible in the domain are with respect to the autonomous car. These actions are \textit{Accelerate}, \textit{Decelerate}, \textit{LeftSqueeze}, \textit{RightSqueeze}, \textit{LeftLight \{On, Off\}}, \textit{RightLight \{On, Off\}}, \textit{SlowDown} and \textit{WaitAtStopSign}. To change a lane, three consecutive actions of \textit{\{Left, Right\} Squeeze} are required. 

From $\mathcal{M^R}$ we generated a total of 40 plans (consisting of both explicable and inexplicable behaviors) for 16 different planning problems. These plans were assessed by 20 human subjects, with each subject evaluating 8 plans (apart from 1 subject who evaluated 7 plans). Also, each plan was evaluated by 4 different subjects. The overall number of training samples was 159. 
The test subjects were required to have a state driving license. The subjects were provided with the initial state and goal of the car. After seeing the simulation the plan, they had to record whether they found each action explicable or not. 
The assessment had two parts: one part involved scoring each autonomous car action with 1, if explicable, and 0 otherwise (plan score was calculated as the fraction of actions in the plan that were labeled as explicable); the other part involved answering a questionnaire on the preconditions, effects of the agent actions. It consisted of 8 questions with \emph{yes/no} answers. 
The questions used for constructing the domain and the corresponding answers are provided in Table \ref{table:question}. For each question, the answers with majority of votes were used by the domain modeler to construct $\mathcal{M^R_H}$. The questions with divided opinions (3 and 7) were not included in the models as some found that behavior explicable while some others did not.
The $\mathcal{M^R_H}$ domain consists of the same state predicates but ended up with different action definitions with respect to preconditions, effects, and action-costs. 

\begin{figure*}[tbp!]\centering
	\begin{subfigure}[t]{0.25\textwidth}
        \includegraphics[height=1.4 in]{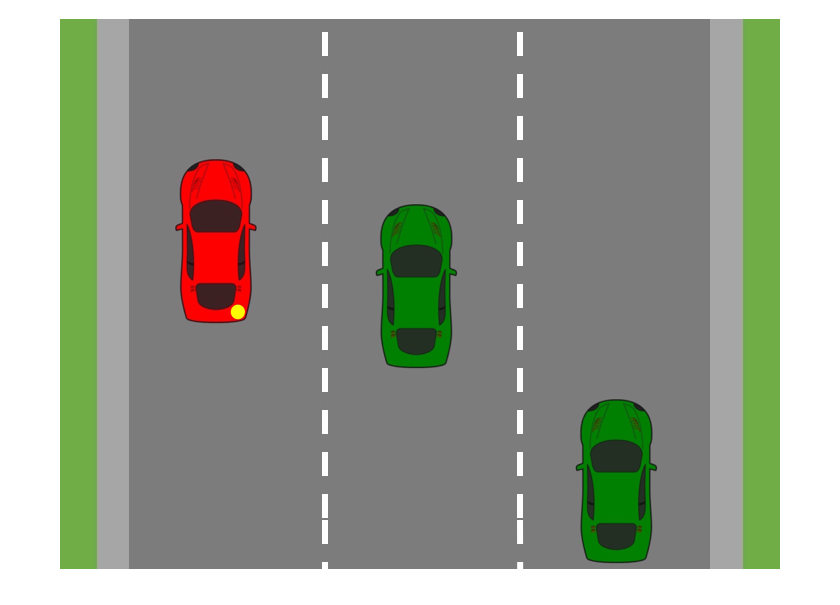}
        \caption{}
        \label{fig:car_sima}
    \end{subfigure}
    ~
    \begin{subfigure}[t]{0.25\textwidth}        \includegraphics[height=1.4 in]{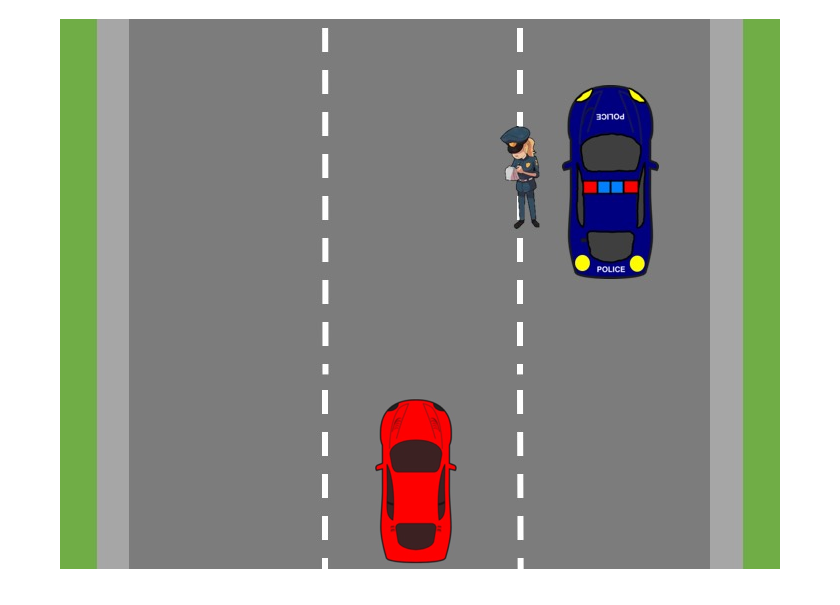}
        \caption{}
        \label{fig:car_simb}
    \end{subfigure}
     ~ 
     \begin{subfigure}[t]{0.25\textwidth}
        \includegraphics[height=1.4 in]{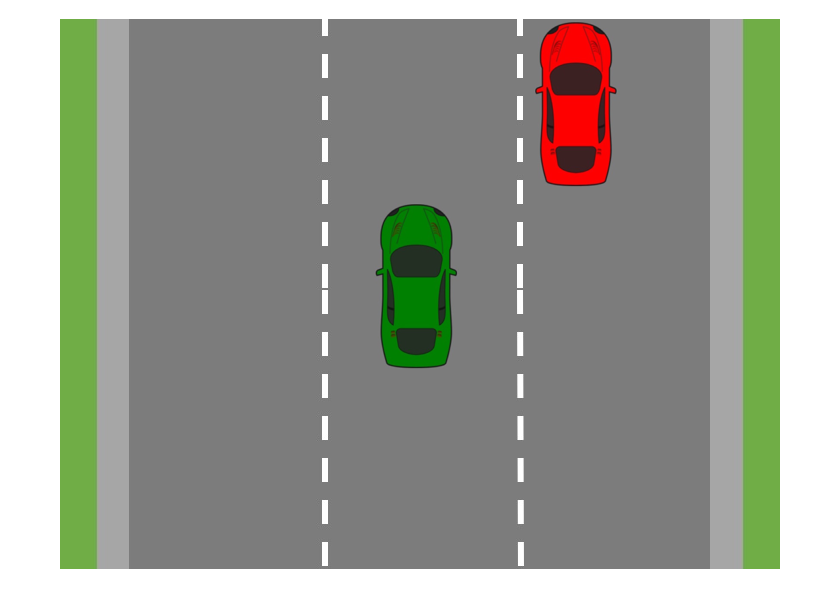}
        \caption{}
        \label{fig:car_simc}
    \end{subfigure}
    ~
    \begin{subfigure}[t]{0.25\textwidth}
        \includegraphics[height=1.4 in]{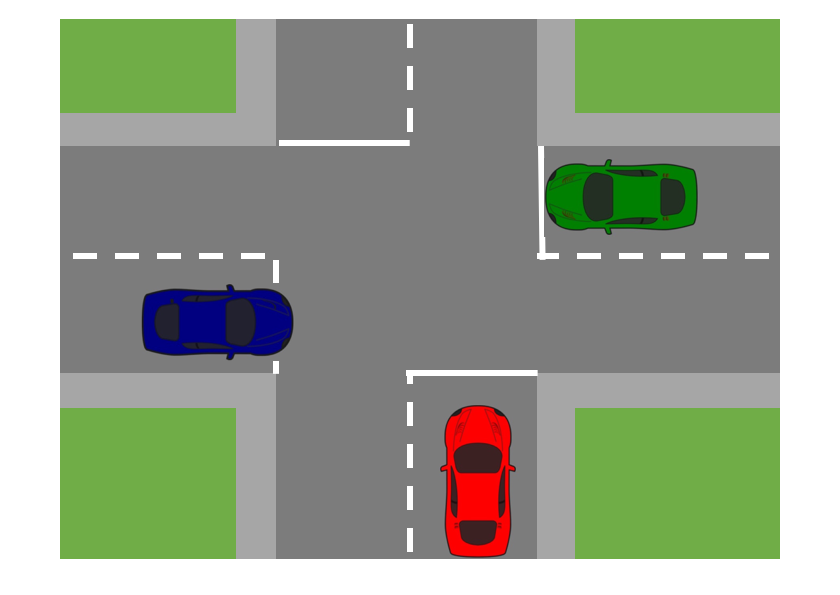}
        \caption{}
        \label{fig:car_simd}
    \end{subfigure}
    \caption{Simulated Autonomous Car Domain. Here only the red car is autonomous. (a) The autonomous car is performing lane change task (b) The autonomous car is performing a move-over maneuver (c) The autonomous car is trying to merge to the middle lane and is confusing the human driver with its signals. (d) The autonomous car is waiting at a 4-way stop even though it is its turn to cross.}
    \label{fig:car_sim}
\end{figure*}

\setlength{\tabcolsep}{3pt}
\renewcommand{\arraystretch}{1.5}
\begin{table}[t!]
\begin{center}
\resizebox{\columnwidth}{!}{%
\begin{tabular}{ |c|C{5.5cm}|c|c| }
\hline
No. & Questions & Yes & No \\
\hline
\hline
1 & Autonomous   car   should   always   maintain   itself   in   the   center   of   the   lane   unless   it   is
changing   lanes & 16 & 4\\
\hline
2 & The car should automatically turn signal lights on before lane change. & 20 & 0 \\
\hline
3 & The car should automatically turn signal lights on when   the   car   starts   swerving
away from the center of the lane. & 9 & 11 \\
\hline
4 & At   a   four-way   stop,   it   should   automatically   provide   turn   signal   lights   when   it   is
taking left or right turns. & 18 & 2 \\
\hline
5 & It should slow down automatically when it is swerving off the center of the lane. & 15 & 5 \\
\hline
6 & It   should   slow   down   automatically   when   there   is   an   obstacle   (pedestrian   or parked   vehicle)   ahead   in   the   lane.& 17 & 3 \\
\hline
7 &Check   one: When an emergency vehicle is parked on the rightmost lane, it should (1)  automatically follow the move over maneuver, (2) whenever possible follow the move over maneuver. & 9 & 11 \\
\hline
8 & Check one: At a four-way stop, it should (1) wait for the intersection to be clear and to be extra safe. (2)  wait for the other cars unless it is its turn to cross over. & 15 & 5 \\
\hline

\end{tabular}%
}
\vspace{5pt}
\caption{The questionnaire used in the human study for Car Domain, and the tally of answers given by participants. For the last two questions, the participants were asked to choose one of the two options, and the ``yes" tally corresponds to the first answer, ``no" to the second. }
\label{table:question}
\end{center}
\end{table}

\setlength{\tabcolsep}{2pt}
\renewcommand{\arraystretch}{1.0}
\begin{table}[t!]\tiny
\begin{center}
\resizebox{\columnwidth}{!}{%
\begin{tabular}{ |c|c|c| }
\hline
Algorithm & Car $R^2$ \% & Delivery $R^2$  \% \\
\hline
\hline
Ridge Regression  &53.66 & 31.42\\
\hline
AdaBoost Regression & 61.31 & 66.99 \\
\hline
Decision Tree Regression &74.79 & 39.61\\
\hline
Random Forest Regression & 90.45 &  75.29\\
\hline
\end{tabular}%
}
\vspace{5pt}
\caption{Accuracy for car and delivery domain.}
\label{table:regression}
\end{center}
\end{table}

\begin{figure}[t]
\centering
\includegraphics[width=\columnwidth]{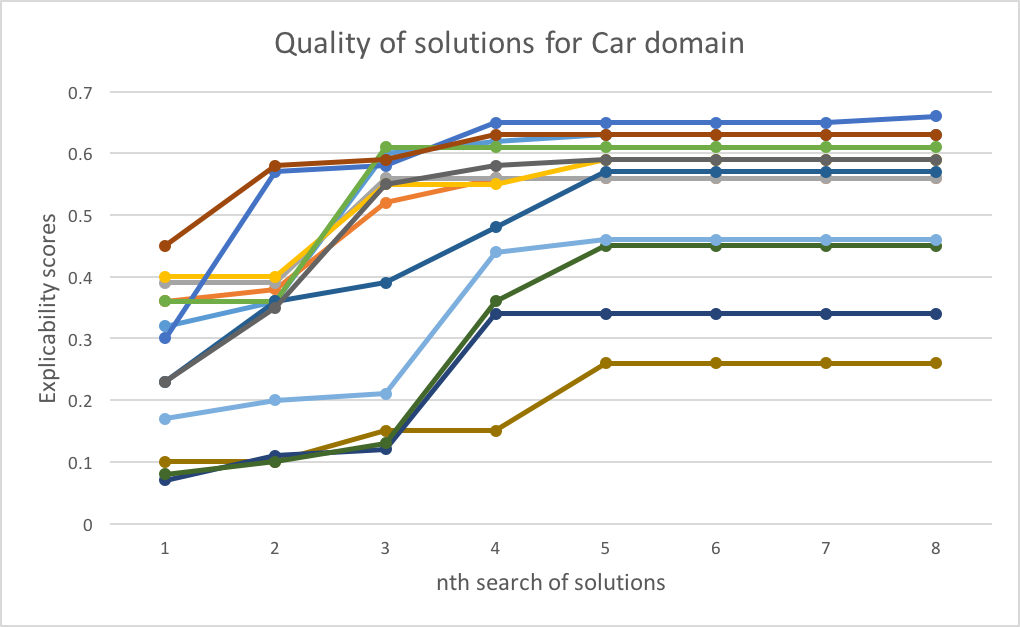}
\caption{For the car domain test problems, the graph shows how the search process finds plans with incrementally better explicable scores. Each color line represents one of the 13 different test problems. The markers on the lines represent a plan solution for that problem. The y-axis and the x-axis represents the explicability scores of the plans and the solution number respectively.}
\label{fig:graph1}
\end{figure}

\iffalse
\begin{figure}[tbp!]\centering
    \begin{subfigure}[t]{0.48\columnwidth}        \includegraphics[height=1.2 in]{car3}
        \caption{}
        \label{fig:car_sima}
    \end{subfigure}
     ~ 
     \begin{subfigure}[t]{0.48\columnwidth}
        \includegraphics[height=1.2 in]{car4}
        \caption{}
        \label{fig:car_simc}
    \end{subfigure}
    \caption{Simulated Autonomous Car Domain. Here only the red car is autonomous. (a) It is having difficulty merging to the middle lane and is confusing the human driver with its signals. (b) It is waiting at a 4-way stop even though it is its turn to cross.}
    \label{fig:car_sim}
\end{figure}
\fi

Following are two examples of how the feedback was interpreted by the domain modeler. For question 1, the majority of answers agreed with the statement.   
Therefore for actions   Accelerate    and   SlowDown,  additional  preconditions  like   \textit{(not (squeezingLeft   ?x)),} \textit{(not   (squeezingLeft2   ?x)),}  \textit{(not (squeezingRight   ?x)), }  \textit{(not   (squeezingRight2   ?x)) }    were  added, where x stands for car.
For   question   8, since   the   answers   agreed   with   choice   2,   actions \textit{waitAtStopSign1},  \allowbreak  \textit{waitAtStopSign2},   \textit{waitAtStopSign3}    were  replaced  by  a new  general  action   $\textit{waitAtStopSign}$.    This  action  removed  predicates      $\textit{waiting1, waiting2, waiting3}$   from  the  action  definition.  Also,     actions \textit{atStopSignAccelerate},   \textit{atStopSignLeft}, \textit{atStopSignRight}    were changed  to  remove  the  precondition   \textit{waiting3 }   (these actions thus had two  definitions in $\mathcal{M^R}$ to allow for explicable behavior, one with higher cost).

\begin{figure}[t]
\centering
\includegraphics[width=\columnwidth]{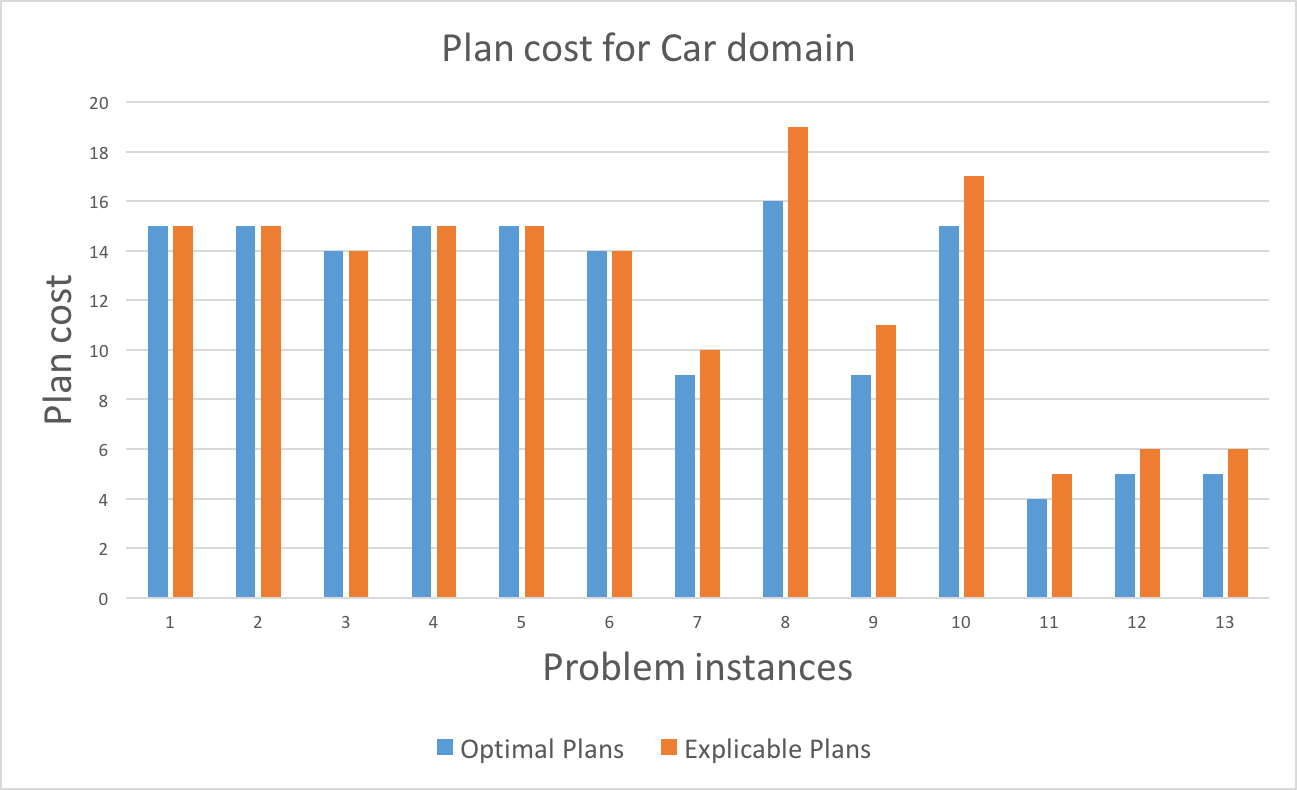}
\caption{For the car domain, the optimal and explicable plans were compared for their explicability scores.}
\label{fig:graph2}
\end{figure}

\begin{figure}[t]
\centering
\includegraphics[width=\columnwidth]{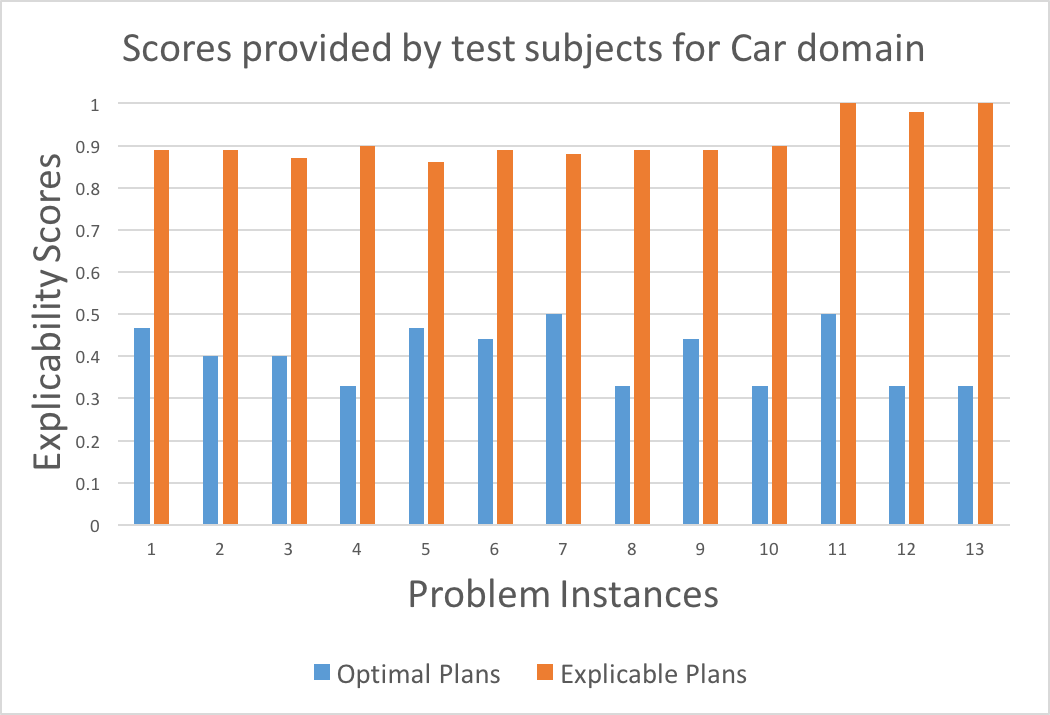}
\caption{For the car domain, the optimal and explicable plans were compared for their explicability scores.}
\label{fig:graph3}
\end{figure}

\iffalse
\begin{figure*}[tbp!]
    \begin{subfigure}[p]{0.33\textwidth}
        \includegraphics[height=1.4 in]{graph1}
        \caption{}
        \label{fig:graph1}
    \end{subfigure}
     ~ 
    \begin{subfigure}[p]{0.33\textwidth}
        \includegraphics[height=1.4 in]{graph2}
        \caption{}
        \label{fig:graph2}
    \end{subfigure}
     ~ 
    \begin{subfigure}[p]{0.33\textwidth}
        \includegraphics[height=1.4 in]{graph3}
        \caption{}
        \label{fig:graph3}
    \end{subfigure}
    \caption{ Car domain: (a) The graph shows how the search process finds plans with incrementally better explicable scores. Each color line represents one of the 13 different test problems. The markers on the lines represent a plan solution for that problem. The y-axis gives the explicability scores of the plans and the x-axis gives the solution number. The optimal plans generated using the \texttt{Fast-Downward} planner were compared for (b) plan costs (c) explicability scores provided by test subjects. 
    }
    \label{fig:results1}
\end{figure*}
\fi

\subsubsection*{Defining the explicability distance}

For the training problems, explicable plans were generated using the model $\mathcal{M^R_H}$. Since some actions names were not common to both the domains, an explicit mapping was defined between the actions over the two domains. This mapping was done in order to support plan distance operations performed between plans in the two domains (for the plan distances to be used effectively common action names are required). 

As noted in Definition \ref{def:4}, features of the regression model are the three plan distances and the target is the score associated with the plans. We tune the hyperparameters by performing grid search over parameters like the number of trees, depth of tree and the minimum number of nodes to perform sample split. The results for different learning models are as shown in Table \ref{table:regression}. We tried several ensemble learning algorithms to improve the accuracy of our model, out of which random forest regression gave the best performance. Random forests allow selection of a random subset of features while splitting the decision node. We evaluate the goodness of fit of the model, using the coefficient of determination or  $R^2$. 
%This value determines the measure by which the fitted model can explain the variations in the target values. This value lies between 0 to 1. Higher the $R^2$ value, better is the model fitted to the data. 
After training process, the new regression model was found to have 0.9045 $R^2$ value. That is to say, 90\% of the variations in the features can be explained by our model. Our model predicts the explicability distance between the agent plans and human mental model plans, with a high accuracy.

\iffalse
\begin{figure}[t]
    \begin{subfigure}[t]{0.45\columnwidth}
        \includegraphics[height=1 in]{graph4}
        \caption{}
        \label{fig:graph4}
    \end{subfigure}
     ~ 
     \begin{subfigure}[t]{0.54\columnwidth}
        \includegraphics[height=1 in]{graph5}
        \caption{}
        \label{fig:graph5}
    \end{subfigure}
    \caption{For delivery domain test problem instances, the optimal and explicable plans were compared for  (a) plan costs (b) explicability scores provided by test subjects.}
\label{fig:Rplot}
\end{figure}
\fi

\subsubsection*{Evaluation}

We evaluated our approach on 13 different planning problems. We ran the algorithm with a high cost bound, in order to cover the most explicable candidate plans for all the problems. The Figure \ref{fig:graph1} reports the explicability scores of the first 8 solutions generated by our algorithm for 13 test problems. From this graph, we note that the reconciliation search is able to develop plans with incrementally better explicability scores. These are the internal explicability scores (produced by the explicability distance function). From Figure \ref{fig:graph2}, we see for the last 7 problems, our planner generated explicable plans with a cost higher than that of optimal plans; a planner insensitive to explicability would not have been able to find these expensive but explicable plans. This additional cost can be seen as the price the agent pays to make its behavior explicable to the human. For the first 6 problems, even though the cost is same as that of the optimal plans, a planner insensitive to explicability cannot guarantee the generation of explicable plans. From Figure \ref{fig:graph3}, the test subjects provided higher explicability scores for the explicable plans than the optimal plans for all 13 problems. 
%For the first 6 problems, the optimal cost planner did not generate explicable plans even though the explicable plans were optimal. 
The plan scores given by 10 test subjects were computed as the ratio of explicable actions in the plans to the total plan length. Testing phase protocol was same as that of the training phase, except for the questionnaire. The plan scores were averaged over the number of test subjects. In this domain, an inexplicable plan for changing lanes from \textit{l2} to \textit{l1} can be \textit{LeftSqueeze-l2-l1, LeftSqueeze-l2-l1, LeftSqueeze-l2-l1, LeftLightOn} whereas an explicable plan would be \textit{LeftLightOn, LeftSqueeze-l2-l1, LeftSqueeze-l2-l1, LeftSqueeze-l2-l1}. The ordering of \textit{LeftLightOn} action decides whether the plan is explicable.

%Also, our system works better against an optimal planner, that is to say, if explicable plans need to be generated the optimal planner will not always produce a plan reasonable with respect to the human mental model of the agent. 

\subsection{Robot based Delivery Domain}

\begin{figure*}[tbp!]
\begin{center}
\includegraphics[width=0.98\textwidth]{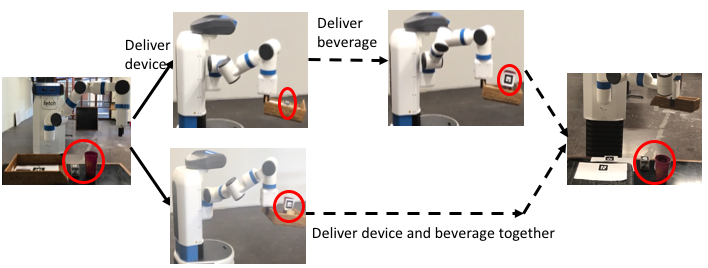}
\caption{The goal of the agent is to deliver the device and beverage cup to the destination. In the cost-optimal plan, agent delivers both the items together, whereas in the explicable plan the agent delivers the items separately. A video demonstration can be viewed at \protect\url{https://bit.ly/2JweeYk}}
\label{fig:trace}
\end{center}
\end{figure*}

This domain is designed to demonstrate inexplicable behaviors of a delivery robot.
The robot can deliver parcels/electronic devices and serve beverages to the humans. It has access to three regions namely kitchen, reception and employee desk. For the evaluation, we used a Fetch robot, which has a single arm with parallel grippers for grasping objects. It delivers beverages, parcels, and devices using a tray. Whenever the robot carries the beverage cup there is some risk that the cup may tip over and spill the contents all over the electronic items on the tray. Here the robot has to learn the context of carrying devices and beverages separately even if it results in an expensive plan (in terms of cost or time) for it.
A sample plan in this domain with explicable and inexplicable plans is illustrated in Figure \ref{fig:trace}. Here, in the inexplicable version, the robot delivers device and beverage together. Although it optimizes the plan cost, the robot may tip the beverage over the device. Whereas, in the explicable version robot delivers the device and beverage cup separately, resulting in an expensive plan due to multiple trips back and forth. 
An anonymous video demonstration can be viewed at \protect\url{https://bit.ly/2JweeYk}.

\begin{figure}[t]
\centering
\includegraphics[width=\columnwidth]{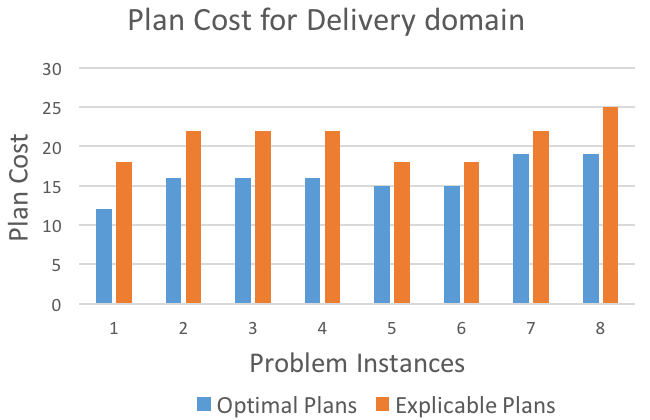}
\caption{Comparison of plans costs for the optimal and explicable plans in the delivery domain.}
\label{fig:graph4}
\end{figure}

\begin{figure}[t]
\centering
\includegraphics[width=\columnwidth]{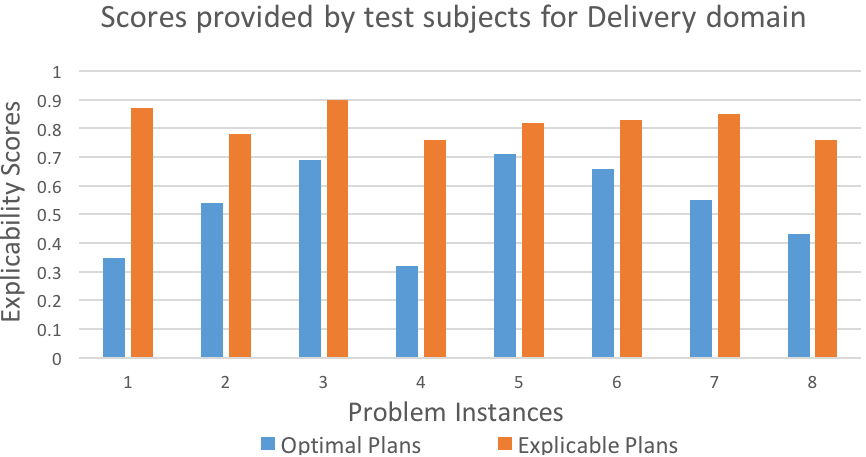}
\caption{Comparison of the explicability scores provided by the test subjects for the optimal and explicable plans in the delivery domain.}
\label{fig:graph5}
\end{figure}

\paragraph{Domain and explicability distance}

This domain is also represented in PDDL. Here both the models were provided by the domain expert. The robot model has the following actions available: \texttt{pickup}, \texttt{putdown}, \texttt{stack}, \texttt{unstack} and \texttt{move}. The domain modeler provided $\mathcal{M^R_H}$ based on usual expectations of robots with a similar form factor. For example, in $\mathcal{M^R_H}$, certain actions which could be perceived riskier (like, carrying the device and cup in the same tray) had a higher cost due to the possibility of damaging the items. Thus $\mathcal{M^R_H}$ incentivizes the planner to choose safer actions. Both models have same state space and action representation. The model differences lie in the action-costs as well as preconditions and effects of actions. There were 20 plan instances created for each of the 13 planning problems. Each of the plans was labeled by 2 human subjects, which resulted in 40 different training samples (some problems have multiple solution plans). The performance of different ensemble learning techniques is as shown in Table \ref{table:regression}. We again use the random forest regression model with an accuracy of 75\%. 

\paragraph{Evaluation}
For evaluation of this domain, 8 new planning problems (similar to the one shown in Figure \ref{fig:trace}) were used. These plan instances with a pictorial representation of intermediate behavioral states were labeled by 9 test subjects. For testing phase, the same protocol was followed as that in training phase. Figure \ref{fig:graph4} shows the comparison between the plan costs. Whenever the items consist of beverage cups, the robot has to do multiple trips, therefore all the explicable plans are more expensive than the optimal plans. 
For such scenarios, if a robot uses a cost-optimal planner, the plans chosen will always be inexplicable with respect to the plan context. In Figure \ref{fig:graph5}, we compare the explicability scores provided by test subjects. The explicability scores provided by the subjects are higher for explicable plans. Some plans involved the robot stacking cups over devices to generate cost-optimal plans. These plans ended up receiving least scores.

\section{Related Work}

An important requirement for an AI agent collaborating with a human is the ability to infer human's goals and intents and use this information to inform its planning process. There have been various efforts in the direction of human aware planning to account for human's beliefs and goals \cite{sisbot2007human,cirillo2010human,mainprice2011planning,serendipity,chakraborti2016planning} to encourage human agent interactions. Human-AI collaboration can involve both purely virtual tasks \cite{allen1994mixed,ai2004mapgen,sengupta2017radar}, purely physical tasks \cite{alami2005task,montreuil2007planning} or their combination. 

HRI community has focused on some related issues in the context of the human-AI collaboration, but they have mainly considered purely physical tasks involving manipulation and motion. Our multi-model explicability framework can be adapted to the HRI scenarios too. Indeed, there has been some work on explicability of motion planning that goes under the name of \emph{predictability} \cite{Dragan-RSS-13,predictability,knepper2017implicit}. Predictability assumes the goal is known and deals with generating likely (straightforward) motion towards it (when the goal is unknown, legibility is defined in this context, which conveys with a motion which goal the robot is reaching for). These can be understood in terms of $\mathcal{M^R}$ - $\mathcal{M^R_H}$ models. The difference is that for motion tasks, the advantage is that humans have well understood intuitive/common-sense models (e.g. move in straight lines as much as possible). This reduces the difficulty of generating explicable plans considerably, as the human model is fixed, and the \emph{distance} between the expected and observed trajectories can be easily measured. However, in many general domains that may involve both physical and virtual tasks (or purely virtual tasks), there are no such widely accepted intuitive human models. In such cases, the full generality of our approach will be necessary.
%\footnote{Note that even in the case of physical robots, all behavior cannot be reduced to motion/manipulation alone. For example, consider a robot that makes a straight-line motion towards its goal, while emitting an ear-piercing sound on the way. In such a case, while the motion itself is predictable by definitions of \cite{Dragan-RSS-13}, the overall behavior may well be inexplicable.}

Previously, \cite{exp-yz} showed the generation of explicable plans by approximating $\mathcal{M^R_H}$ to a labeling scheme given that was learned using conditional random fields. In this work, we study how the problem changes when the human model is known and is represented in terms of an action model similar in representation to the agent's model. In many domains (household/factory scenarios), there is a clear expectation of how a task should be performed and $M^R_H$ can be constructed using relevant norms. Most deployed products make use of inbuilt models of user expectations. This allows for a more informed plan generation process, which we refer to as \emph{reconciliation search}. This assumption on model representation also allows us to investigate the relationship between the traditional measures of plan distances studied in the planning community and \emph{explicability measure} implicitly used by the humans to compare the agent plans with the expected plans from their model. 

This work also has connections with the work on generating explanations through ``model reconciliation'' \cite{explain} (not to be confused with ``reconciliation search'' introduced in this paper). The generation of explicable plans and explanations can be seen as complementary strategies to deal with model differences. In \citet{explain}, the authors make a similar assumption of having access to the human mental model. The major difference is that in the work on explanations, in order to explain an inexplicable plan the model differences are communicated to the human in order to update their incomplete or noisy model. However, in our work, we do not communicate the model differences, instead generate plans that are closer to the human's expected plan thereby making the plan more explicable than it was before. The act of providing explanations can increase the human's cognitive load especially when the human is operating in a mission critical scenario or the communicated explanation involves multiple updates. In such situations, the computation of explicable plans would be more suitable and would not lead to increase in the cognitive load as long the explicable plan is closer to the expected plan. In certain domains, communication constraints can prevent communication of explanations altogether. In such situations, generating explicable behavior can be the only alternative available to the robot.

\section{Conclusion}

In conclusion, we showed how the problem of explicable plan generation can be modeled in terms of plan distance measures studied in existing literature. 
We also demonstrated how a regression model on these distance measures can be learned from human feedback and used to guide an agent's plan generation process to produce increasingly explicable plans in the course of an anytime search process.  
We demonstrated the effectiveness of this approach in a simulated car domain and a physical robot delivery domain. We reported the results of user studies for both the domains which show that the algorithmically computed explicable plans have a higher explicability score than the optimal plans.

Going forward, two interesting avenues of future work include (1) learning to model plan similarities when domains are represented at different levels of abstraction (e.g. with experts versus non-experts in the loop); and (2) exploring interesting modes of behavior that can build upon the developed understanding of human expectations (e.g. by being inexplicable to draw attention to an event or, in the extreme case, being deceptive by conforming to or reinforcing wrong human expectations). The former may require a rethink of existing plan distance metrics to account for the effects of model difference in the presence of lossy abstractions.

\section*{Acknowledgments} 

This research is supported in part by the ONR grants N00014-16-1-2892, N00014-18-1-2442, N00014-18-1-2840, the AFOSR grant FA9550-18-1-0067, and the NASA grant NNX17AD06G.

\bibliographystyle{ACM-Reference-Format}
\balance
\bibliography{bib}

\end{document}